# ComplexFace: a Multi-Representation Approach for Image Classification with Small Dataset

*Guiying Zhang[a,b,#], Yuxin Cui[b,#] , Yong Zhao[c], And Jianjun Hu[b,*]*

[a] School of Basic Medical Sciences, Guangzhou Medical University, Guangzhou, Guangdong Province, 511400, PR China;
[b] Department of Computer Science and Engineering, University of South Carolina, Columbia, South Carolina, 29208, USA;
[c]The Key Lab of Integrated Microsystems, Peking University Shenzhen Graduate School, Shenzhen, Guangdong, 518055, PR China

\# Equal contribution
\* *jianjunh@cse.sc.edu*

**Abstract：** State-of-the-art face recognition algorithms are able to achieve good performance when sufficient training images are provided. Unfortunately, the number of facial images is limited in some real face recognition applications. In this paper, we propose ComplexFace, a novel and effective algorithm for face recognition with limited samples using complex number based data augmentation. The algorithm first generates new representations from original samples and then fuse both into complex numbers, which avoids the difficulty of weight setting in other fusion approaches. A test sample can then be expressed by the linear combination of all the training samples, which mapped the sample to the new representation space for classification by the kernel function. The collaborative representation based classifier is then built to make predictions. Extensive experiments on the Georgia Tech (GT) face database and the ORL face database show that our algorithm significantly outperforms existing methods: the average errors of previous approaches ranging from 31.66% to 41.75% are reduced to 14.54% over the GT database; the average errors of previous approaches ranging from 5.21% to 10.99% are reduced to 1.67% over the ORL database. In other words, our algorithm has decreased the average errors by up to 84.80% on the ORL database.

## 1. Introduction

Face recognition is one of the most active research topics in computer vision[1, 2]. During the past several years, a large number of algorithms have been proposed for face recognition [3, 4], for which extraction of effective features is most critical. Traditional feature transformations such as principal component analysis (PCA)[5, 6], Fisher discriminant analysis (FDA)[7] and Locality Preserving Projection (LPP)[8], are widely used for feature manipulation in face recognition. Later, those features are replaced by local features such as Local Binary Patterns (LBP)[9, 10], Local Phase Quantisation (LPQ)[11], as well as Scale-invariant feature transform (SIFT)[12]. For datasets with simple background, traditional features can achieve very competitive results. Unfortunately, for uncontrolled environment, the face recognition performance with such traditional features may degrade dramatically. It is challenging to improve the performance of face recognition because facial images are affected by variation in illumination, facial expression and poses.

In recent years, Convolutional Neural Networks (CNNs) have achieved great success in computer vision such as objection detection, objection recognition, image segmentation, action recognition, and especially face recognition [13-15]. There are two main characteristics that make CNNs achieve such outstanding success. First, parallel GPUs greatly improve the training efficiency of CNNs. Second, a lot of labelled training data are available to learn rich features by the training algorithm of CNNs, which can automatically learn the features of facial images by tuning the large number of weights. However, CNN requires large-scale training datasets with labels. This means that highly efficient hardware and large-scale annotated data are the key to obtain the high performance of CNNs. This is, however, not available in applications in which there is no GPU for the target hardware or the number of training samples are limited, which leads to overfitting for CNNs. In such cases, non-CNN methods are needed to achieve competitive performance of face recognition. Among these approaches, generating new representations from original images and fusing multiple representations is a simple and effective method to improve the robustness of face recognition.

Data augmentation via generating new presentations from original images with mathematical functions can enlarge the training datasets and can achieve higher classification accuracy. These approaches can mainly be categorized into two categories. In the first type of approaches, new representations are obtained via the face structure, including symmetrical characteristic of faces and the mirror image of face [16-19]. In the second category, new representations are constructed by the perturbation and distribution of original samples[20-22]. Different from these two categories of approaches, Xu et al.[23] developed a novel method to automatically produce approximately axis-symmetrical virtual face images. Zhang et al.[24] proposed a method to operate small regions of the original image to obtain new representations. Moreover, singular value decomposition has also been applied to generate new representations in [25, 26]. Both sparse representation based classification[27, 28] and collaborative representation based classification (CRC) [29] have been used with an assumption that a test sample is represented as a linear combination of an entire set of training samples and plays an important role in face recognition[23] [30-32]. They can achieve good performance if there are sufficient training samples for a given class object. Based on this fact, we design our method to obtain new representation from original sample to enlarge the face dataset.

Data sample fusion can offer more supplementary information from multiple samples or information sources and obtain better performance in image classification, which has been shown in many existing methods [18, 19, 24-26, 33, 34]. In these methods, it is a good and effective way that new representations from original samples are generated via related arithmetic operators and then new representations and original images are fused by appropriate weights. Traditional fusion methods have a drawback that the weights for multiple samples need to be manually set via empirical knowledge [24, 25, 35]. To address this problem, we exploit a method to transform the original image and the new representation into a complex number image. This scheme can be regarded as two kinds of images are fused or two features are fused without weight setting. Thus, our method avoids the parameter setting while more complementary information can be available for face recognition.

In addition to complex numbers have been used to represent samples, quaternions are also exploited in pattern classification [36]. A quaternion is composed of four real numbers, so some mathematical operations of a quaternion can be decomposed as operations of two complex numbers. As higher-order extensions of real numbers, complex numbers and quaternions not only allow more information to be integrated into a denotation, but also can exploit crossbred information of original numbers. In this sense, the corresponding method can be viewed as a feature level fusion method.

In existing methods, multi-sample fusion only works in the 2D space so that the supplementary information obtained by multiple samples is limited. If the fusion is in the 3D space, more supplementary information will be available. Motivated by the fact that kernel based methods [37] can be applied to represent and classify test samples by using all the training samples in the feature space, we present a novel method to transform different complex samples including test complex images and training complex images into a new space. This process can be viewed as fusion of two kinds of samples in the 3D space. We apply collaborative representation based classification [29]to conduct the face recognition experiments

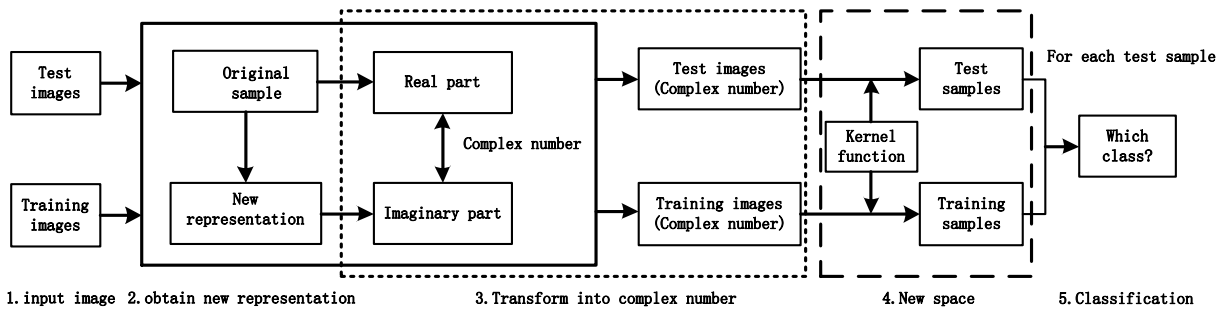

Fig. 1. Overview of our ComplexFace algorithm: it (1) takes input images (test images or training images), (2) obtain new representations from original samples, (3) compose the complex images by taking the original sample as the real part of the complex number (image), and taking the new representation as the imaginary part of the complex number (image), (4) transform the test complex images and training complex images into a new space, respectively, and then (5) classifies each test sample by using collaborative representation based classification (CRC) in the new space.

In this paper, we proposed a novel algorithm, namely, ComplexFace, for face recognition with small datasets, as shown as Fig. 1. Our algorithm uses the data augmentation strategy to generate new representations from original images, and then fuse these two sets of images by applying the complex numbers approach--the original sample as the real part of a complex number and the new representation as the imaginary part of the complex number, which allows to combine the pair-wise images for complementary information. This avoids parameter-setting problem typical for other fusing approaches, which makes it convenient in real-world applications. We assume that the test sample can be expressed by a linear combination of all the training samples, then we apply kernel function to complex numbers images and transform them into a new space. This scheme enables two kinds of samples or features to be fused in a 3D space, thus, more complementary information can be available for more accurate face recognition. Collaborative representation based classification CRC is then built in the new space. Extensive experiments on the GT dataset and the ORL dataset show that the proposed approach outperforms other facial recognition methods, including collaborative representation based classification (CRC) [29], an improvement to the nearest neighbor classifier (INNC)[38], ADWF[26] , KRBM[39] , SVD fusion[25] and method in[24]. For further performance comparisons, we conducted our experiments in comparison to derived images[20] and symmetrical faces [18]. All the experimental results show that our method achieved impressive accuracy rates for face recognition with limited samples.

The remainder of this paper is organized as follows. Section 2 discusses related work. Section 3 presents the proposed approach in detail. Section 4 discusses the experimental results. Section 5 concludes the paper.

## 2. Related Work

Data augmentation by generating new presentations from original samples has been widely used to improve classification accuracy. Passalis et al. [16] proposed a novel face recognition method to handle pose variations by using facial symmetry. Mirror images or symmetrical faces of original faces are applied as new presentations in Xu et al. [17, 18, 40], which achieves favourable face recognition performance. Liu et al. [19] proposed a framework that fused virtual mirror synthesized training samples as bases and hierarchical multi-scale local binary patterns (LBP) features for face classification. Zhang et al. [41] exploited a new face recognition method based on SVD perturbation for single-example image per person. Shan et al. [22] proposed a sample expansion method to generate virtual samples. Xu et al. [23] developed a novel method to automatically produce approximately axis-symmetrical virtual face images. Zhang et al. [24] proposed a novel method to represent an image by first operating the non-overlapping blocks with certain arithmetic, and then merging modified blocks to obtain new representations of original images. Singular value decomposition is applied to generate new representations in [25, 26].

Several sample fusion methods have been proposed to improve face recognition performance. Xu et al. [18] integrated both original images and symmetrical faces for face recognition. This method may result in ugly virtual samples due to the use of the symmetrical faces'. Liu et al. [19] fused hierarchical multi-scale local binary patterns and virtual mirror samples to perform face recognition. Zhang et al [26] proposed a simple and effective method that automatically fuses the original samples and two groups of virtual samples obtained by singular value decomposition. Werghi et al. [42] presented a novel approach to fuse shape and texture local binary patterns (LBPs) on a mesh for 3D face recognition. Ding et al. [43] proposed a new scheme to extract "Multi-Directional Multi-Level Dual-Cross Patterns" (MDML-DCPs) from face images, which outperforms the state-of-the-art local descriptors for both face identification and face verification tasks.

Deep learning has also been exploited to address face recognition task. Deepface [44] took the last hidden layer neuron activations of deep convolutional networks to obtain the features. This study used a common method that augments the dataset via obtaining new samples in a multi-scale and multi-channel way. Following DeepID2 [45], DeepID2+ [46], DeepID3 [47] and method in [48, 49] have been proposed to further improve the performance for face recognition. Schroff et al. proposed a method named FaceNet [50] that applied a deep convolutional network trained to directly optimize the embedding itself, which reduced the error rate in comparison to DeepId2+.

## 3. The Proposed Method

In this section, we describe the different components of our ComplexFace method as shown in Fig.1 in detail. We firstly input facial images. We then discuss the method for generating new representations from original samples. Following we transform both the original samples and the new representations into complex numbers. Afterwards we show how to transform test complex samples and training complex samples into a new space. In the end, the collaborative representation based classification is used to conduct our experiments.

### 3.1 Image transformation

There are $s$ classes among $n$ original training samples represented as $x_k^j \in \Re^{a*b} (k = 1,...,s, j = 1,...,n)$ . $x_k^j$ represents the $j$ -th sample of the $k$ -th class. We define $X = \left[ x_1^1...x_1^n,...,x_s^1...x_s^n \right]$ as the set of all original training samples. In our method, each original training sample $x_k^j$ is converted into size $a*b$ -dimensional row vector, there are $sn$ samples in $X$ , so we treat the original training samples as a matrix of size $sn*ab$ , which is composed by $sn$ rows and $ab$ columns. In other words, the original samples set $X = \left[ x_1^1...x_1^n,...,x_s^1...x_s^n \right]$ can be viewed as a matrix with size $sn*ab$ .

The multi-representations of a sample allow to bring additional supplemental information, which may lead to higher face recognition accuracy. To obtain more representations of the original training sample, we obtain new representations of the original training sample using the following formula

$$tx_k^j = x_k^j \bullet \left( 255 - x_k^j \right), \ \ k = 1,...,s, j = 1,...,n \tag{1}$$

to obtain the new image $tx_k^j$ , where $tx_k^j$ represents the new representation of the original sample, and symbol ' $\bullet$ ' represents dot product. We define $TX = \left[ tx_1^1...tx_1^n,...,tx_s^1...tx_s^n \right]$ as the set of new representations from the original training samples.

In the similar way, we define $y_k^j \in \Re^{a*b} (k = 1,...,s, j = 1,...,m)$ as a test sample, then the test sample set is $Y = \left[ y_1^1...y_1^m,...,y_s^1...y_s^m \right]$ , and $Y$ is a matrix with size by $sm*ab$ . We obtain new representations of the original test sample by using the following formula

$$ty_k^j = y_k^j \bullet \left( 255 - y_k^j \right), \ \ k = 1,...,s, j = 1,...,m \tag{2}$$

$TY = \left[ ty_1^1...ty_1^m,...,ty_s^1...ty_s^m \right]$ is the set of new representations from the original test samples. Fig. 2. shows some of original images and corresponding new representations from the ORL dataset.

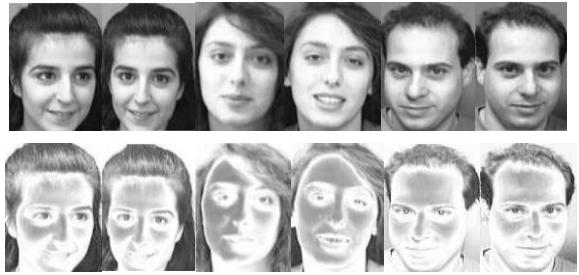

Fig. 2. Sample original images and their corresponding new representations from the ORL dataset. First row are the original images and second row are the corresponding new representations from the ORL dataset。

### 3.2 Integrating the original images and their new representations

A complex number $h = p + qi$ has a real component $p$ and an imaginary component $q$ . For each original image, we compose a complex number by taking the original sample as the real number and the new representation from the original image as the imaginary unit. For each training sample $x_k^j$ and its new representation $tx_k^j$ ,we compose the corresponding complex number as follows

$$cx_k^j = x_k^j + tx_k^j i, \ \ k = 1,...,s, j = 1,...,n \tag{3}$$

$CX = \left[ CX_1,\cdots,CX_s \right] = \left[ cx_1^1...cx_1^n,...,cx_s^1...cx_s^n \right]$ is the set of training complex samples.

Similarly, for each test sample $y_k^j$ and its new representation $ty_k^j$ , we use the below formula to obtain the corresponding complex number

$$cy_k^j = y_k^j + ty_k^j i, \ \ k = 1,...,s, j = 1,...,m \tag{4}$$

$$CY = \begin{bmatrix} CY_1, \cdots, CY_s \end{bmatrix} = \begin{bmatrix} cy_1^1...cy_1^m,...,cy_s^1...cy_s^m \end{bmatrix}$$ is the set of test complex samples.

### 3.3 Transforming the samples into a new space

Suppose $X = \begin{bmatrix} x_1^1...x_1^n,...,x_s^1...x_s^n \end{bmatrix}$ is the original training samples, we obtain the corresponding training complex numbers set $CX$ via equations (1) and (3). In the similar way, for test samples $Y = \begin{bmatrix} y_1^1...y_1^m,...,y_s^1...y_s^m \end{bmatrix}$, we obtain the corresponding test complex numbers set $CY$ via equations (2) and (4) as well. Suppose $z$ is a test sample from test complex samples set $CY$. We apply a nonlinear mapping $\phi$ to the test sample $z$ and training complex samples $CX$. Suppose the test sample $\phi(z)$ can be represented by a linear combination of all the training samples. In other words, we have

$$\phi(z) = \sum_{i=1}^{s} \alpha_i \phi(CX_i) \tag{5}$$

to represent the test sample $z$. We can rewrite formula (5) as following form

$$\phi(z) = \Phi\Gamma \tag{6}$$

where $\Phi = \begin{bmatrix} \phi(CX_1), \cdots, \phi(CX_n) \end{bmatrix}$, $\Gamma = (\alpha_1, \cdots, \alpha_n)^T$. To obtain the solution of formula (6). Both sides of the formula are multiplied by $\Phi^T$, then we have

$$\Phi^T \phi(z) = \Phi^T \Phi\Gamma \tag{7}$$

We apply kernel function $k(X_i, X_j) = \phi^T(X_i)\phi(X_j)$ to work on formula (7), then we have

$$K_z = K\Gamma \tag{8}$$

Where $K_z = \begin{pmatrix} k(X_1, z) \\ \vdots \\ k(X_n, z) \end{pmatrix}$, $K = \begin{pmatrix} k(X_1, X_1) \cdots k(X_1, X_s) \\ \vdots \\ k(X_s, X_1) \cdots k(X_s, X_s) \end{pmatrix}$, $\Gamma = \begin{pmatrix} \alpha_1 \\ \vdots \\ \alpha_n \end{pmatrix}$

### 3.4 Recognition procedure

We suppose that $z \in CY$ is a test complex sample, $CX = \begin{bmatrix} CX_1, \cdots, CX_s \end{bmatrix}$ is the training complex sample set. We define the following objective function via a modified collaborative representation method

$$\| z - CX\theta \|_2^2 + \lambda \| \theta \|_2^2 \tag{9}$$

Which is to be minimized. $\lambda$ is set to a positive constant. The solution to Eq. (9) can be obtained by

$$\alpha = (CX + \lambda I)^{-1} z \tag{10}$$

or

$$\beta = (CX^T CX + \lambda I)^{-1} CX^T z \tag{11}$$

Eq. (11) is known as the solution of collaborative representation.

Based on (9), the matching score of test sample $z$ with regard to the $i$-th class is evaluated by

$$d^i = \| z - CX\theta_i \|_2 \tag{12}$$

where $\theta$ is obtained by Eq. (10) or Eq. (11). In other words, $\theta$ is one of the $\alpha$ and $\beta$.

In short, the main steps of our algorithm are described as follows:

Step 1. For the set of original training sample set $X$ and test sample set $Y$, we apply Eqs. (1) and (2) to obtain new representations $TX$ and $TY$.

Step 2. For new representation $TX$ and $TY$, we obtain training complex samples $CX$ using Eq. (3) and obtain the test complex samples $CY$ by Eq. (4).

Step 3. We apply kernel functions to transform the samples into new space by using Eqs. (5), (6), (7) and (8).

Step 4. Collaborative representation based classification is applied to training samples via Eqs. (9) and (11). Then we calculate the residual error for each training sample by Eq. (12). Finally we classify the test sample $z$ into the $cls$-th class by using $cls = \arg\min_i d_i$.

## 4. Experimental Results

We perform an extensive evaluation of our method on the most widely used and representative benchmark face datasets: the Georgia Tech (GT) face database and the ORL face database. All the experiments demonstrated that the performance of our method is significantly better than that of the compared methods.

### 4.1 Datasets

The ORL face database [51] consists of 40 subjects in total. For each subject, there are 10 images with different ages, genders and nations. Some of the images in ORL have various properties such as pose, expression and decorative features (e.g. glasses or no glasses). At the same time, there is 20% scale change in those images. All the images were acquired against a dark homogeneous background with the subjects in an upright and frontal position. Each image was resized to 46×56 pixels in our experiments. Fig. 3 shows some of facial images in ORL dataset.

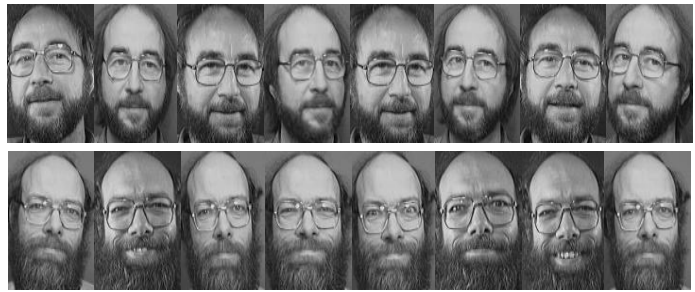

Fig. 3. Sample facial images in ORL dataset

The Georgia Tech (GT) face database is composed of 50 persons, each person has 15 colourful images with clutter background. GT is provided by the Georgia Institute of Technology and is freely available to everyone [52]. The size of the images in GT is 640*480 pixels, we used down-sampled face images of size 30×40 pixels in our experiments. The details for obtaining these images are presented in [17]. Those images have different expressions, various illuminations and diverse scales while including two angles: frontal face and side face. Fig. 4 shows some of facial images in GT dataset.

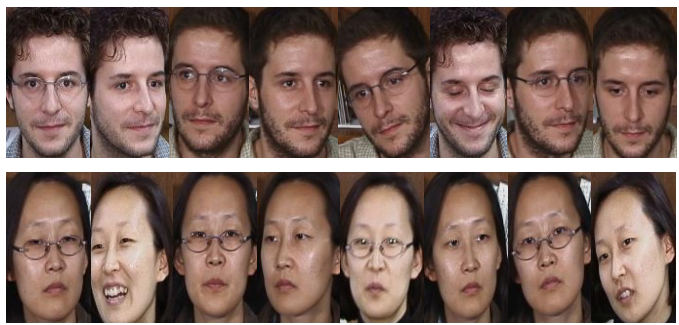

Fig. 4. Sample facial images in GT dataset

### 4.2 Performance analysis for the proposed method

In this section, we describe the extensive experimental results of our method: ComplexFace. In all our experiments, we apply the kernel function $k\left(x_i, x_j\right) = \exp\left[-\left\|x_i - x_j\right\|^2 \middle/ (2\sigma)\right]$ to transform the samples into a new space. The parameter $\lambda$ is set to 0.01. We define $\sigma$ as $10^{10}$ according to the analytical results discussed in detail in Section 4.3. To illustrate the outstanding performance of our algorithm, we compare ComplexFace with the standard collaborative representation based classification (CRC) [29], an improved nearest neighbour classifier (INNC)[38], ADWF [26], KRBM[39], and the method in [24]. All the experiments are described below and ComplexFace shows significant performance improvement over all existing methods.

### 4.2.1 Experiments on the ORL dataset

For each class on the ORL database, we treated first 6 facial images as the training samples and the remain 4 facial images as the test samples. To illustrate how the number of training samples to affect the performance, we also evaluated our method by using the first 7, 8 or 9 facial images as the training samples and the remaining images (3, 2 or 1) as the test samples respectively. Table 1 shows all the experimental results. Just as shown for the GT dataset, our method also greatly outperforms all the other methods over the ORL database. Our method reduces the average classification error rates of with CRC (8.70%) and INNC (8.33%) to 1.67%, which is more than 80% drop in average error rate. This means that ComplexFace algorithm is very useful in improving the accuracy of face classification. It is worth mentioning that when the number of training samples is 8 or 9, both classification error rates are 0, which is an outstanding result.

**Table 1. Experimental results (classification error rates: %) on the ORL face database**

| T_S per class | 6 | 7 | 8 | 9 | Avg_error |
|---|---|---|---|---|---|
| CRC[29] | 6.88 | 9.17 | 11.25 | 7.50 | 8.70 |
| INNC[38] | 8.75 | 10.83 | 6.25 | 7.50 | 8.33 |
| ADWF [26] | 5.00 | 5.83 | 5.00 | 5.00 | 5.21 |
| KRBM [39] | 10.63 | 8.33 | 8.75 | 10.00 | 9.43 |
| SVD fusion[25] | 5.00 | 5.83 | 6.25 | 5.00 | 5.52 |
| Method in [24] | 7.50 | 9.17 | 10.00 | 7.50 | 8.54 |
| **ComplexFace** | **5.00** | **1.67** | **0.00** | **0.00** | **1.67** |

### 4.2.2 Experiments on the GT dataset

On the GT face database, the first 9, 10, 11, 12 or 13 face images of each class is treated as training samples and the remaining images is regarded as test samples. All the comparable experimental results are shown as Table 2, which show that our method greatly outperforms all the other methods. The average error rate (our method is 14.62%) was 18.88% lower compared to that of CRC (33.50%). What's more, the average error rate of our method is 14.62%, which is 27.13% lower compared to that of KRBM (41.75%). Both results demonstrated that ComplexFace algorithm has achieved significant improvement over all existing methods compared here.

**Table 2. Experimental results (classification error rates: %) on the GT face database**

| T_S per class | 9 | 10 | 11 | 12 | 13 | Avg_error |
|---|---|---|---|---|---|---|
| CRC[29] | 39.00 | 36.00 | 32.50 | 32.00 | 28.00 | 33.50 |
| INNC[38] | 38.00 | 36.40 | 33.50 | 30.00 | 26.00 | 32.78 |
| ADWF [26] | 36.00 | 34.80 | 32.50 | 30.00 | 25.00 | 31.66 |
| KRBM[39] | 48.33 | 44.40 | 43.00 | 38.00 | 35.00 | 41.75 |
| SVD fusion[25] | 36.67 | 35.60 | 33.00 | 29.33 | 25.00 | 31.92 |
| Method in[24] | 36.00 | 34.40 | 32.00 | 31.33 | 26.00 | 31.95 |
| **ComplexFace** | **18.67** | **17.20** | **15.50** | **13.33** | **8.00** | **14.54** |

### 4.3 Extended experiments for our method

To further understand the performance of ComplexFace algorithm, we conducted additional performance comparisons and parameter tuning experiments as shown in this section.

### 4.3.1 Comparison with other fusion methods

One of the most important contributions of our method is the use of complex numbers to fuse the new representation and the original image to provide more information for face recognition. To analyse the performance of our fusing method, we conducted experiments to compare with the methods in [18] and [41]. The former method [18] exploited the symmetry of the face to generate new samples. We integrate the symmetrical face obtained by this work and the original image to perform experiments. The latter method [41] was proposed by Zhang et al. In comparative experiments, we set n as 5/4 and $\beta$=0.25 with more details in [41]. The experimental results shown in Table 3 illustrate that our method achieves outstanding results compared to the other methods. It is worth mentioning that we maximally decrease the average errors by 84.80% on the ORL database (The average error by using Derived images + original images is 10.99%, and our method's average error is only 1.67%). That is a very encouraging result.

**Table 3.  Experimental results (classification error rates: %) on the ORL face database**

| T_S per class | 6 | 7 | 8 | 9 | Avg_error |
|---|---|---|---|---|---|
| Symmetrical faces+ original images[18] | 8.75 | 9.17 | 7.50 | 10.00 | 8.86 |
| Derived images + original images[41] | 11.87 | 10.83 | 11.25 | 10.00 | 10.99 |
| Combined images + original images[41] | 10.63 | 8.33 | 10.00 | 7.50 | 9.12 |
| Virtual images + original images[25] | 5.00 | 5.83 | 6.25 | 5.00 | 5.52 |
| **ComplexFace** | **5.00** | **1.67** | **0.00** | **0.00** | **1.67** |

### 4.3.2 Evaluation for the value of coefficients

There are two solutions for eq. (9), one is

$$\alpha = \left( K\left( X,X \right) + \lambda I \right)^{-1} z \tag{10}$$

And the other is

$$\beta = (K\left( X,X \right)^{T} K\left( X,X \right) + \lambda I)^{-1} K\left( X,X \right)^{T} z \tag{11}$$

For further evaluation, we test our method using different parameters for eq. (10) and eq. (11) respectively. Table 4 shows the classification errors on the GT face database. In this table, for example, $\alpha, \sigma = 10^{7}$ represents that we use eq. (10) to obtain parameter $\alpha$ and use the parameter $\sigma = 10^{7}$ in experiments. In the similar way, $\beta, \sigma = 10^{7}$ represents that we use eq. (11) to obtain parameter $\beta$ and use the parameter $\sigma = 10^{7}$ in experiments. If the $\sigma$ is fixed, the classification error using the parameter $\beta$ is a little bit better than the classification error using the parameter $\alpha$ (22.69% vs 22.76%, and 14.54% vs 14.62%).

**Table 4.  Experimental results (classification error rates% ) on the GT face database**

| T_S per class | 9 | 10 | 11 | 12 | 13 | Avg_error |
|---|---|---|---|---|---|---|
| $\alpha, \sigma = 10^{7}$ | 27.67 | 26.80 | 25.00 | 21.33 | 13.00 | 22.76 |
| $\beta, \sigma = 10^{7}$ | 27.67 | 26.80 | 25.00 | 21.00 | 13.00 | 22.69 |
| $\alpha, \sigma = 10^{10}$ | 19.33 | 17.60 | 15.50 | 12.67 | 8.00 | 14.62 |
| $\beta, \sigma = 10^{10}$ | 18.67 | 17.20 | 15.50 | 13.33 | 8.00 | 14.54 |

For the ORL dataset, similar to the GT database, we repeated our experiments via the four situations as shown in Table 5. No matter what the values we set for the coefficients, we can obtain the same classification error rates. That means our method is robust with low dependency on the coefficient parameters of the algorithm.

**Table 5. Experimental results (classification error rates: %) on the ORL face database**

| T_S per class | 6 | 7 | 8 | 9 | Avg_error |
|---|---|---|---|---|---|
| $\alpha, \sigma = 10^{7}$ | 13.30 | 7.50 | 10.00 | 7.50 | 9.58 |
| $\beta, \sigma = 10^{7}$ | 13.30 | 7.50 | 10.00 | 7.50 | 9.58 |
| $\alpha, \sigma = 10^{10}$ | 5.00 | 1.67 | 0.00 | 0.00 | 1.67 |
| $\beta, \sigma = 10^{10}$ | 5.00 | 1.67 | 0.00 | 0.00 | 1.67 |

### 4.3.3 Setting algorithm parameter $\sigma$

To show how parameter $\sigma$ in the eq. (8) affects the prediction performance, we conducted an experiment for different training samples and different databases. We set $\alpha = \left( K\left( X, X \right) + \lambda I \right)^{-1} z$ in this experiment. The experimental results on the GT database is shown as Fig. 2. The horizontal coordinate represents the exponent for parameter $\sigma$, for example, the value of 10 in the horizontal coordinate means that the value of parameter $\sigma$ is $10^{10}$, in the similar way, the value 11 means and the $\sigma$ is $10^{11}$. The vertical coordinate represents the error rates obtained by different training numbers and different values of the parameter $\sigma$ in Fig. 5.

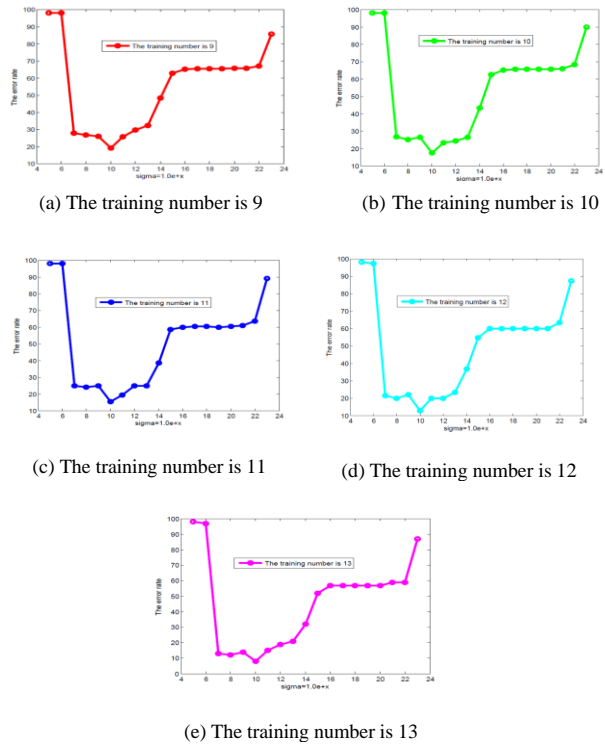

(a) The training number is 9

(b) The training number is 10

(c) The training number is 11

(d) The training number is 12

(e) The training number is 13

Fig.5 The error rate (%) for different $\sigma$ on the GT database: (a) represents the error rate for different $\sigma$ ($\sigma=10^4, \sigma=10^5, \cdots, \sigma=10^{24}$) when training number is 9 and test number is 6. In a similar way, b, c, d, e represents the error rate for different $\sigma$ ($\sigma=10^4, \sigma=10^5, \cdots, \sigma=10^{24}$) when training number is 10, 11, 12, 13, and corresponding test number is 5, 4, 3, 2, respectively.

Fig.5 (a) represents the error rates obtained when the training number is 9 for each class and the remaining samples are the test samples. Fig.5 (b), (c), (d) and (e) have the same meaning as Fig.5 (a). As shown as Fig.5, all curves reached the minimum error rate if $\sigma=10^{10}$. According to this fact, we set $\sigma=10^{10}$ to conduct all our experiments.

To further evaluate parameter $\sigma$, we also conducted an experiment on the GT database. The results are shown in Fig. 6, The horizontal coordinate represents the exponent for parameter $\sigma$, and the vertical coordinate represents that the error rate obtained by different training number and different value of the parameter $\sigma$. Again, all the curve has the minimum error rate if the $\sigma=10^{10}$.

## 5. Conclusions

Data augmentation by fusing raw images with transformed images is helpful for improving the performance of face recognition algorithms. In this paper, we proposed ComplexFace, a novel algorithm for face recognition with limited training samples. It first generates new representations from the original images and the fuse both into complex numbers by taking the new representation as the imaginary part of the complex number and the original sample as the real part of the complex number. Then we apply kernel functions to transform complex number samples into a new representation space. Our experimental results of this collaborative representation based classification showed that our fusion approach is able to achieve significant performance improvement compared to the state-of-the-art face recognition methods for limited samples, reducing their error rates by up to 84.80% on the ORL dataset. The source code of our program will be provided upon request.

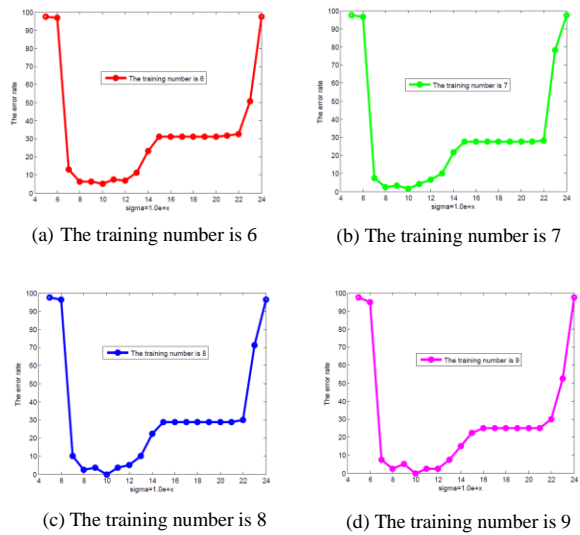

(a) The training number is 6      (b) The training number is 7

(c) The training number is 8      (d) The training number is 9

Fig.6 The error rate (%) for different $\sigma$ on the ORL database. (a) represents the error rate for different $\sigma$ ( $\sigma=10^4, \sigma=10^5, \cdots, \sigma=10^{24}$ ) when training number is 6. In a similar way, b, c, d represents the error rate or different $\sigma$ ( $\sigma=10^4, \sigma=10^5, \cdots, \sigma=10^{24}$ ) when training number is 7, 8, 9, and corresponding test number is 3, 2, 1, respectively.

## Acknowledgments

No potential conflict of interest was reported by the authors.